\documentclass[10pt,twocolumn,letterpaper]{article}

\usepackage[pagenumbers]{wacv} 

\usepackage[dvipsnames]{xcolor}

\usepackage{times}
\usepackage{epsfig}
\usepackage{graphicx}
\usepackage{amsmath}
\usepackage{amssymb}
\usepackage{multirow}
\usepackage{xhfill}

\usepackage{url}            
\usepackage{booktabs}       
\usepackage{amsfonts}       
\usepackage{nicefrac}       
\usepackage{microtype}      

\usepackage{changepage}
\usepackage{subcaption}
\usepackage{scrextend}
\usepackage{makecell}
\usepackage{paralist} 
\usepackage{float}
\usepackage{indentfirst}
\usepackage{longtable}
\usepackage{caption}
\usepackage{enumitem}
\usepackage{refcount}

\usepackage{afterpage}

\usepackage[pagebackref,breaklinks,colorlinks]{hyperref}

\usepackage{indentfirst}
\usepackage{longtable}
\usepackage{caption}
\usepackage[accsupp]{axessibility}  

\newcommand{\fusecap}{\textsc{FuseCap}}

\usepackage[capitalize]{cleveref}
\crefname{section}{Sec.}{Secs.}
\Crefname{section}{Section}{Sections}
\Crefname{table}{Table}{Tables}
\crefname{table}{Tab.}{Tabs.}

\def\wacvPaperID{1604} 
\def\confName{WACV}
\def\confYear{2024}

\begin{document}

\title{\fusecap{}: Leveraging Large Language Models \\ for Enriched Fused Image Captions}

\author{
Noam Rotstein* \qquad David Bensa\"id*  \qquad Shaked Brody \qquad Roy Ganz \qquad Ron Kimmel 
 \\
  Technion - Israel Institute of Technology
 \\
\tt\small *Indicates equal contribution.
}
\maketitle

\begin{abstract}
The advent of vision-language pre-training techniques enhanced substantial progress in the development of models for image captioning.
However, these models frequently produce generic captions and may omit semantically important image details.
This limitation can be traced back to the image-text datasets; while their captions typically offer a general description of image content, they frequently omit salient details.
Considering the magnitude of these datasets, manual reannotation is impractical, emphasizing the need for an automated approach.
To address this challenge, we leverage existing captions and explore augmenting them with visual details using ``frozen'' vision experts including an object detector, an attribute recognizer, and an Optical Character Recognizer (OCR).
Our proposed method, \fusecap{}, fuses the outputs of such vision experts with the original captions using a large language model (LLM), yielding comprehensive image descriptions.
We automatically curate a training set of 12M image-enriched caption pairs.
These pairs undergo extensive evaluation through both quantitative and qualitative analyses.
Subsequently, this data is utilized to train a captioning generation BLIP-based model.
This model outperforms current state-of-the-art approaches, producing more precise and detailed descriptions, demonstrating the effectiveness of the proposed data-centric approach.
We release this large-scale dataset of enriched image-caption pairs for the community.
\end{abstract}




\begin{figure}
\centering
\includegraphics[width=\linewidth]{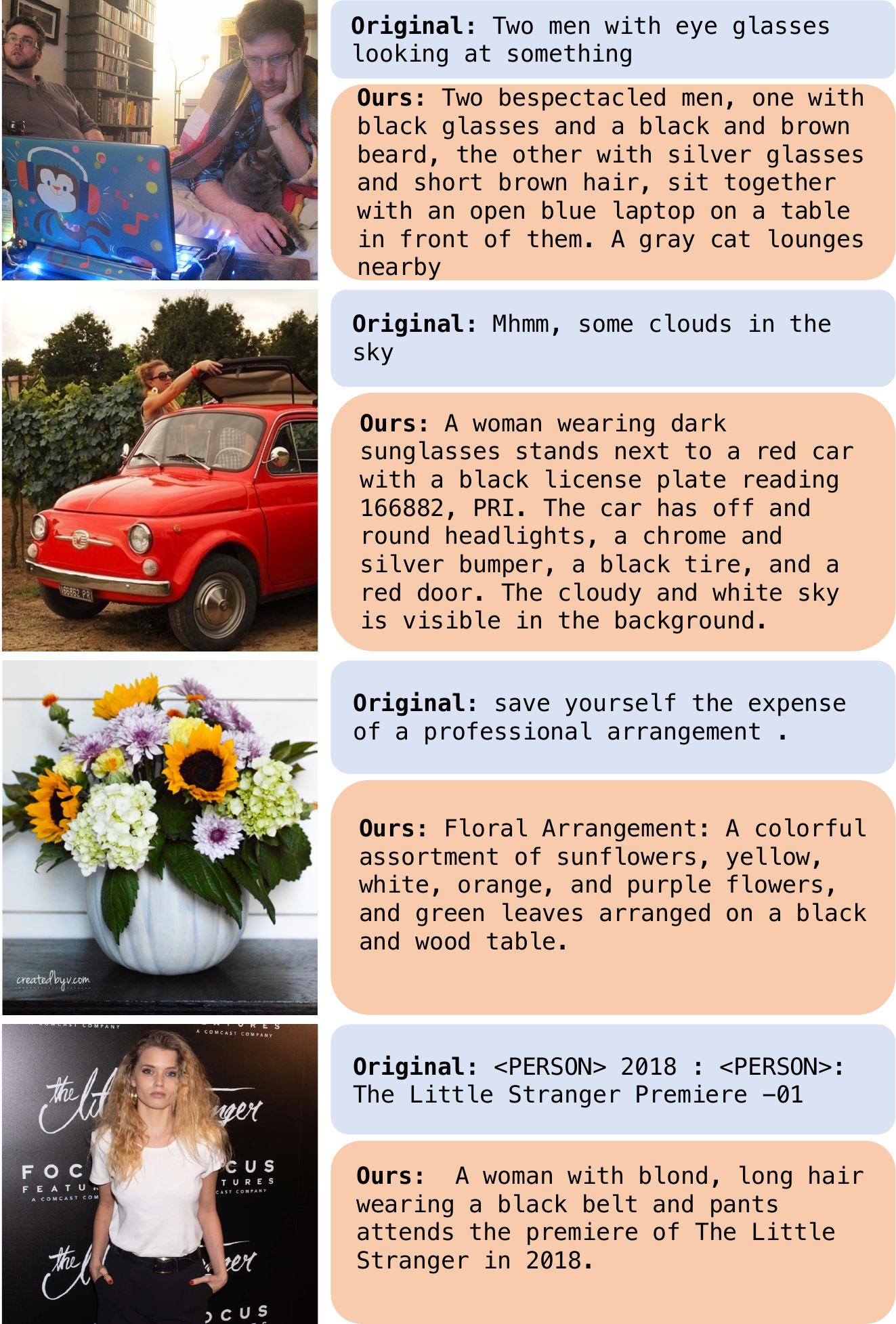}
\caption{\textbf{\fusecap{} captions.} An illustration comparing our \fusecap{} enriched captions with the original ground-truth captions before the fusing process.
The examples are from COCO, SBU, CC, and CC12 datasets, displayed from top to bottom.
}
\label{fig:generatior_example2}
\vspace{-0.46cm}
\end{figure}


\section{Introduction}\label{Se:intro}
The generation of image captions that effectively capture essential descriptive elements has been a longstanding goal in computer vision~\cite{longstanding0,longstanding1,longstanding2,longstanding3,longstanding4,longstanding5}.
In recent years, image captioning tasks~\cite{lin2014microsoft,agrawal2019nocaps} have gained significant research attention and interest due to the success of Vision Language (VL) models.
This achievement mainly stems from the ability to efficiently harness the massive amount of image-caption pairs accessible online, using Vision Language Pre-training (VLP)~\cite{align_before_fuse,uniter,radford2021learning}, followed by task-specific fine-tuning.
However, despite remarkable advancements in image captioning, current state-of-the-art models ~\cite{vinvl,simvlm,li2022blip,li2023blip,hu2022scaling, wang2022git,pali,wang2022ofa} produce captions that often overlook key semantic elements.
As images are rich sources of information containing intricate and complex content, providing precise descriptions requires highly detailed textual captions.

We hypothesize that the current unsatisfactory captioning results are attributed to the image-caption datasets used for training.
Captions in these datasets frequently fail to capture essential elements within images and often omit fine details.
For example, consider the original caption of the top image in \Cref{fig:generatior_example2}.
The caption is missing details such as the laptop and cat.
Since these datasets contain a massive number of image-caption pairs, manual re-annotation is unfeasible.

In this work, our primary objective is to develop a framework that produces richer, more accurate captions for images.
To achieve this, we introduce \fusecap{}, a novel approach designed to automatically augment captions in existing datasets, thereby enhancing training data of the model.
This contrasts with methods that primarily focus on improving the caption generator model architecture and resonates with the recently surveyed data-centric artificial intelligence (AI) paradigm \cite{zha2023data, zha2023data2}, which underscores the significance of improving both the quality and quantity of data, rather than merely concentrating on the advancement of model design.
We leverage the capabilities of \textit{vision experts} such as object detectors \cite{anderson2018bottom}, attribute recognizers \cite{zhang2021vinvl}, and Optical Character Recognition (OCR) models \cite{bautista2022parseq,aberdam2022multimodal,aberdam2023clipter}.
The visual information extracted by these models is intended to provide complementary details to the original simplistic captions.
By harnessing the reasoning capabilities of a dedicated LLM, the outputs from the vision experts are fused with the original image caption.
This results in a coherent, meaningful natural description of images that is more comprehensive and detailed than the original caption, as illustrated in \Cref{fig:generatior_example2}.
Specifically, we leverage ChatGPT\cite{brown2020language} to generate ``fusing'' examples which are then used to fine-tune a pre-trained Flan-T5 model \cite{raffel2020exploring}.
We apply this method to enrich captions of a human-annotated dataset (COCO~\cite{lin2014microsoft}) and large-scale datasets collected from the web (CC12~\cite{changpinyo2021conceptual}, CC~\cite{sharma2018conceptual}, and SBU~\cite{vicente2016large}).
This process produces an enriched collection consisting of 12M image-text pairs. 
To confirm the quality of our generated dataset, we first show that humans favor the fused captions, perceiving them as more descriptive and accurate than the original captions.
Furthermore, we demonstrate that the fused captions score 
higher on CLIPScore~\cite{hessel2022clipscore} -- a reference-free metric that evaluates text-image alignment without using reference captions — compared to the original ones.
To further emphasize their effectiveness, we also assess the captions through image-to-text and text-to-image retrieval tasks.

To illustrate the benefits of the proposed data-centric approach we capitalize on these fused captions to train a caption generator.
We use the augmented dataset both for pre-training and for the fine-tuning of an image captioning BLIP model \cite{li2022blip}.
Despite having fewer parameters and using less training data, our model surpasses existing state-of-the-art methods \cite{li2022blip,li2023blip,wang2022ofa,git, liu2023prismer} in generating comprehensive captions.
This superiority is evident both in its higher CLIPScore and its improved performance in the retrieval tasks.
The performance advantage is further illustrated by numerous examples.

\textbf{Our contributions are as follows:}

\begin{itemize}
    \item Introducing \fusecap{}- a novel approach to automatically enrich existing image-captions datasets by fusing outputs from visual experts using an LLM.
    \item Providing a large dataset of 12M caption-enriched text-image pairs for future research.
    \item Showcasing that an enriched dataset leads to models capable of generating detailed captions that effectively incorporate previously overlooked key semantic elements. 
\end{itemize}
\vspace{-0.5cm}

\begin{figure*}[!t]
\centering
\begin{subfigure}{\textwidth}
  \centering
  \includegraphics[width=0.87\linewidth]{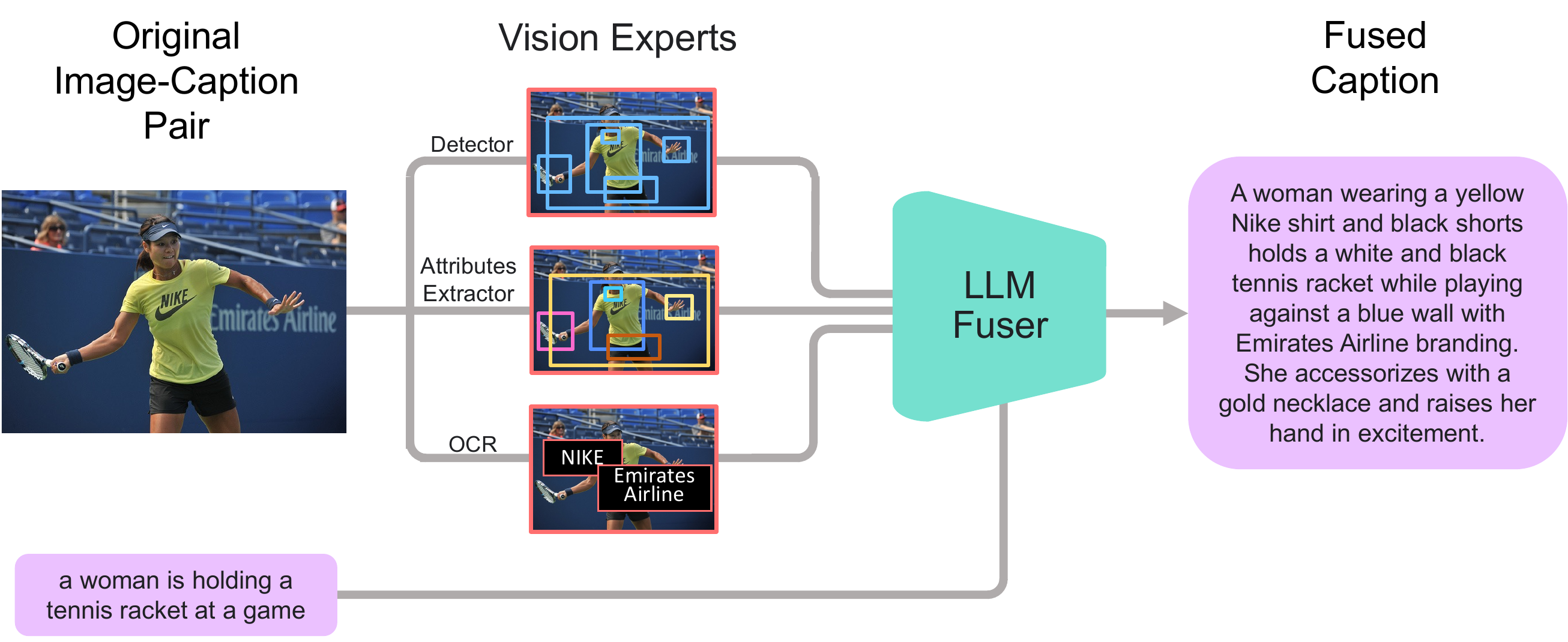}
  \caption{Fusing Enriched Captions}
  \label{method_stage1}
\end{subfigure}
\begin{subfigure}{\textwidth}
  \centering
  \includegraphics[width=0.85\linewidth]{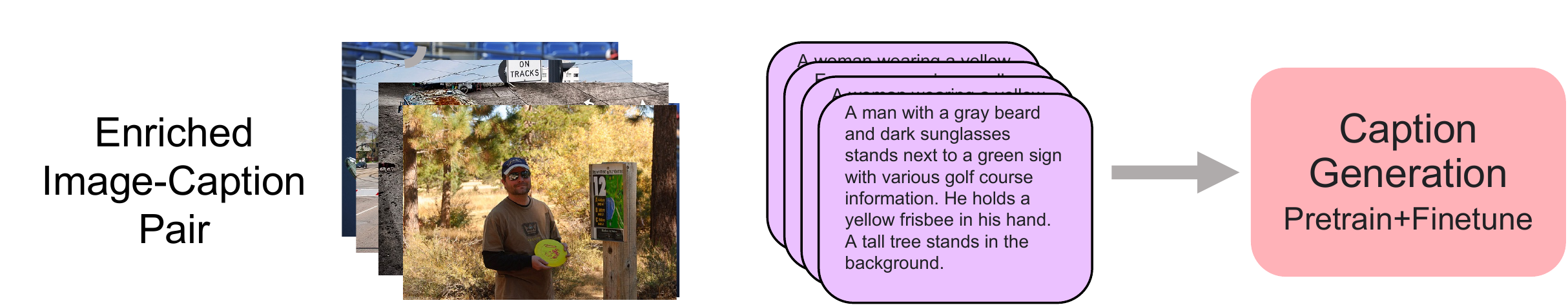}
  \caption{Training a Captioning Model}
  \label{method_stage2}
\end{subfigure}
\caption{
\textbf{Our Approach Illustration.}
\Cref{method_stage1} illustrates the automated process of enriching existing image captions using the proposed \fusecap{} approach.
Visual experts extract meaningful information from images, which is then fused with the original captions by an LLM Fuser, producing rich captions.
Following this, \Cref{method_stage2} illustrates the utilization of image datasets, paired with these augmented captions, in both the pre-training and fine-tuning phases of a comprehensive image-captioning model.
}
\vspace{-0.21cm}
\label{fig:test}
\end{figure*}

\section{Related Efforts}\label{Se:related}

\paragraph{Image-Caption Generation.}
Image captioning has been a widely researched topic at the intersection of computer vision and natural language processing. 
Early strategies for caption generation made use of retrieval-based methods \cite{farhadi2010every, hodosh2013framing} and template-based methods \cite{kulkarni2013babytalk}, which were limited in their expressiveness \cite{bai2018survey}. 
The advent of deep learning marked a shift in this field, as multimodal neural networks enabled the generation of higher quality captions \cite{karpathy2015deep}.
Subsequently, the encoder-decoder framework, which essentially translates an image into a sentence, became one of the favored approaches \cite{vinyals2015show, donahue2015long}.
Later advancements incorporated attention mechanisms for focusing on key image aspects \cite{xu2015show, yang2016review}.
Current best captioning techniques are transformer-based architectures \cite{vaswani2017attention}, which combine vision \cite{dosovitskiy2020image}  and language \cite{vaswani2017attention} transformers.
The recent advancements in the effectiveness of image captioning can largely be credited to the introduction of {\it visual language pre-training} (VLP).

VLP  uses large-scale image-text pairs to pre-train vision-text models, which are later fine-tuned for downstream tasks, with image captioning as one of the central tasks  \cite{li2022blip,li2023blip,ganz2023towards}.
VLP can be categorized \cite{chen2023vlp} into single-stream \cite{li2019visualbert, chen2020uniter} and dual-stream architectures \cite{dou2022empirical, radford2021learning, jia2021scaling,ganz2023clipag}, aiming to merge visual and textual modalities into a shared embedding space.
Efforts exploring integrating text generative tasks for pre-training \cite{wang2021simvlm, cho2021unifying} have fueled the development of models like BLIP \cite{li2022blip} and recent BLIP-2 \cite{li2023blip}.
OFA \cite{wang2022ofa} proposed unifying multiple unimodal and multimodal pre-training tasks, which led to a significant performance improvement.

Dense captioning can be seen as a task related to ours in generating comprehensive text for images \cite{johnson2016densecap, yang2017dense}.
However, it produces multiple captions for various regions within an image, as opposed to a single descriptive caption for the whole image.
Previous approaches to generating more comprehensive captions have consistently identified a common challenge -- the unsatisfactory capability of existing image captioners of providing detailed and accurate textual descriptions of images~\cite{liu2018show, luo2018discriminability, liu2019generating}.
These approaches focus on improving captions discriminability, whereas our method seeks to enrich captions with additional, meaningful information extracted from the images.
\vspace{-6pt}

\paragraph{Image-Caption Datasets}
Current datasets for image-caption pairs fall into two primary categories: specifically human-annotated datasets like COCO \cite{lin2014microsoft}, and web-crawled datasets such as CC, CC12, and SBU Captions \cite{sharma2018conceptual, changpinyo2021conceptual, ordonez2011im2text}.
The conjunction of the datasets serves as the foundation for the VLP, subsequently followed by downstream tasks fine-tuning.
In general, the first category of datasets is smaller in size yet exhibits substantially lower noise levels compared to the second category.
Both categories, however, are characterized by relatively short and concise captions.
Specific examples of these characteristics can be seen in \Cref{fig:generatior_example2}.

Data-centric AI underscores the importance of refining data quality to boost model performance rather than fine-tuning model designs.
It prioritizes curating, labeling, and cleaning data for superior training datasets \cite{sun2017revisiting} as well as automating its processing \cite{zha2023data, zha2023data2}.
This new emphasis suggests that with high-quality data, even basic algorithms can deliver impressive results.
\cite{shi2021enhancing} identified the need to enrich image-text datasets.
To this end, they harnessed natural language inference (NLI) to fuse multiple existing ground-truth captions into a single one. 
However, this approach can operate only on datasets with multiple ground-truth captions and cannot be applied to large-scale ones with a single caption (\textit{e.g.,} CC, CC12, and SBU).
In contrast, our method can be applied to any image-caption dataset.
\vspace{-6pt}

\paragraph{Large Language Models}
Large language models (LLMs) have been shown to be effective in a wide range of tasks, including natural language inference, question answering, and code generation \cite{chung2022scaling, wei2021finetuned, chowdhery2022palm}.
Further, LLMs such as GPT-3 \cite{brown2020language} exhibit impressive zero and few-shot performance on a variety of tasks, including translation, text summarization, and common sense reasoning, without further fine-tuning.
Few-shot abilities allow researchers to use LLMs as a tool for data generation.
\cite{brooks2022instructpix2pix} used GPT-3 to generate instructions and edited captions dataset, which is then used to train model for image editing.
\cite{schick2023toolformer} presented an approach for training LLMs to use external API calls. For example, when the LLM is been asked to solve a mathematical problem, it can use an API of a calculator, instead of generating the output by itself. \cite{schick2023toolformer} used GPT-3 few-shot ability to curate a dataset of external API calls which is then been used to fine-tune another LLM. 
\cite{peng2023instruction} used GPT-4 outputs to create an instruction-following dataset that can be used later to fine-tune other LLMs in a supervised learning fashion.
To fuse together the original caption and the visual expert outputs, in this work, we harness the impressive zero-shot capabilities of OpenAI's ChatGPT \cite{brown2020language}.
We employ it to generate a small ``fusing'' dataset. To establish an open-source framework that scales cost-effectively, we fine-tune Flan-T5 \cite{chung2022scaling}, a widely recognized open-source LLM, using this data.
\vspace{-6pt}

\paragraph{Vision Experts in VLP}
Several works attempted to improve VLP by incorporating object detectors or other experts \cite{liu2023prismer} as part of their initialization \cite{li2019visualbert, cho2021unifying}, architecture \cite{liu2023prismer}, pre-training data \cite{wang2022position} or pre-training objectives \cite{tan2019lxmert, su2019vl, zhang2021vinvl, li2020oscar}.
In image captioning, such models demonstrate limited capabilities in generating rich captions and have not fully capitalized on the information provided by vision experts. We hypothesize that this limitation stems directly from the succinct captions in existing image-text datasets.
In our data-centric approach, we focus on improving such datasets and overcoming this limitation.
An example that highlights our approach is the contrast with Prismer \cite{liu2023prismer}. Unlike our method, Prismer augments its model architecture with object detectors and OCR for caption generation but relies on traditional captioning datasets.


A recent concurrent paper \cite{wang2023caption} proposed an LLM-based model that merges captions of image segments into text.
In contrast, we adopt the LLM from a data-centric perspective to enhance existing caption datasets, an approach that is followed by extensive evaluations for quality and consistency.
\section{Fusing Enriched Captions}\label{Se:method}


In this section, we introduce \fusecap{}, which is illustrated in \Cref{method_stage1}.
This novel strategy is designed to automatically augment existing image captions by integrating important details and inter-object relations within the image.
These details are often disregarded in traditional image captioning datasets.
First, we elaborate on our use of pre-trained vision models, referred to as \textit{vision experts}, for extracting relevant visual information from images.
We then detail how the information gathered from these expert models is subsequently fused with the original caption through a fine-tuned LLM, resulting in the enhanced captions.

\subsection{Vision Expert Models}\label{Se:vision-experts}
To enrich the information found in the original caption, we employ the following vision experts:
\vspace{-8pt}

\paragraph{Object Detection}

A key visual expert we rely on is an object detector model.
Following the approach proposed in 
\cite{zhang2021vinvl}, we utilize a Faster-RCNN \cite{ren2015faster} with a ResNeXt152 \cite{xie2017aggregated} backbone.
This model is initially pre-trained on several detection datasets and then fine-tuned on the Visual Genome (VG) dataset \cite{krishna2017visual}.
The data used comprises more than 100K images with 1.6K classes, enabling strong generalization abilities. 
We regard all objects along with their corresponding bounding boxes as valid detections, provided they exceed a detection confidence threshold.
The presence and position of objects in the scene provide essential details to complement the caption.
\vspace{-8pt}

\paragraph{Attributes Prediction}
Beyond class identification, we derive a variety of attributes for each object in the image using features generated by our Faster-RCNN within each bounding box.
A classification model proposed in \cite{zhang2021vinvl} was trained on annotations from the VG dataset, which covers a broad array of 400 distinct attributes.
These attributes encapsulate various aspects of the objects, including size, condition, and color.
For each object, we only consider attributes predicted with a confidence level above the threshold.
\vspace{-8pt}

\paragraph{Text Detection and Recognition}
Text within images often contains critical contextual information, leading to a line of different text-image tasks \cite{biten2019scene, ganz2023towards}.
To incorporate textual information, we utilize robust pre-trained OCR models to detect and extract characters.
We first identify text within a scene with CRAFT \cite{baek2019character}, a robust scene-text detector.
We then apply Parseq \cite{bautista2022parseq},  a state-of-the-art scene text recognizer, to decode the text within the bounding boxes generated by the text detector.
To avoid contamination, we do not apply OCR methods on datasets presenting watermarks, such as the CC \cite{sharma2018conceptual} and CC12 \cite{changpinyo2021conceptual}.

\begin{figure*}[!htb]
\centering
  \includegraphics[width=1\linewidth]{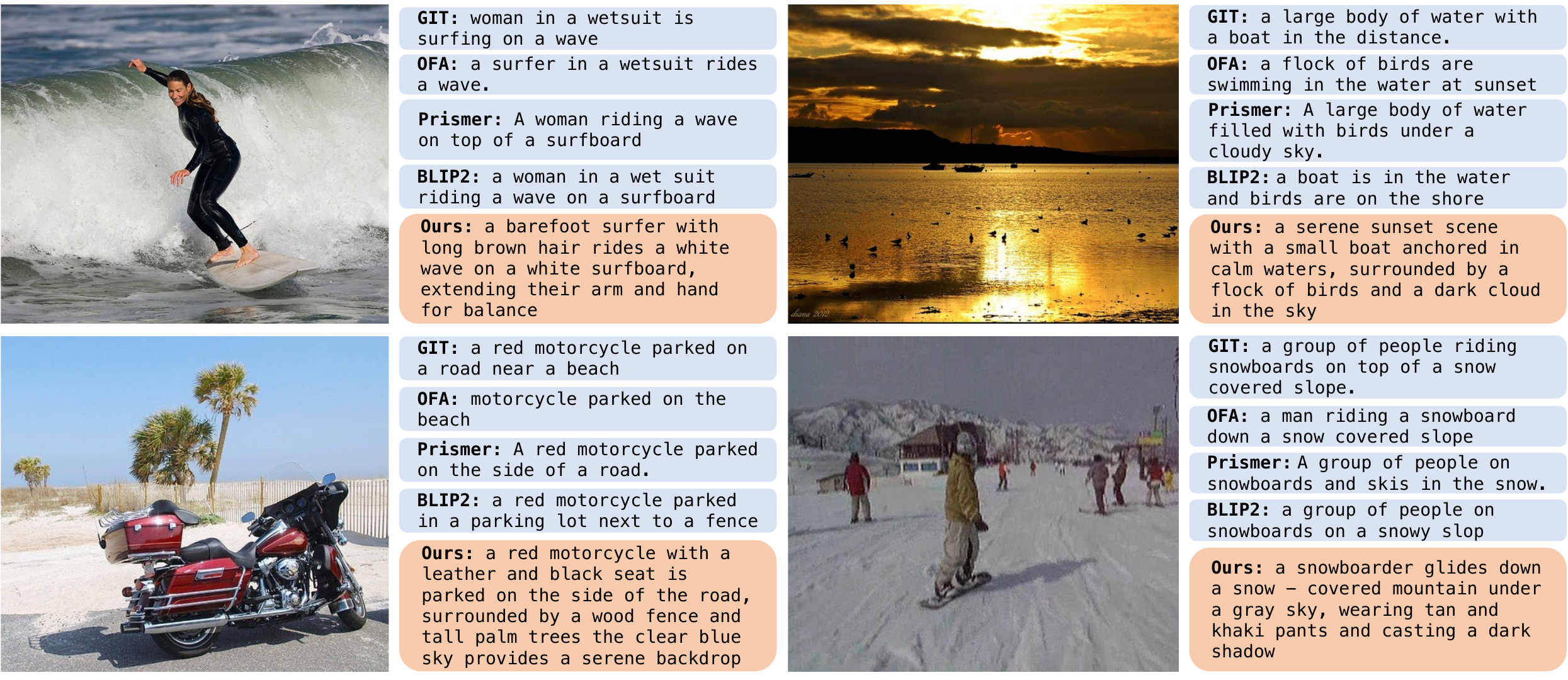}
\vspace{-12pt}
\caption{\textbf{Image Captioning Results.} While top-performing captioning models tend to provide concise and oversimplified captions, our model outputs rich captions that better describe the images.}
\label{fig:generatior_example}
\vspace{-3pt}
\end{figure*}

\subsection{LLM Fuser}\label{Se:llm-fuser}
To generate natural and coherent captions, which are essential for the caption generation task, we utilize a large language model (LLM).
We leverage a specifically fine-tuned LLM to fuse insights from various vision experts into the original caption, creating a single coherent and fluent description.
The LLM's advanced reasoning capabilities allow us to articulate the semantic relationships among objects and to seamlessly integrate the diverse knowledge provided by these experts. 
Consequently, the output captures the essence of the visual content while the impressive generative capacity of LLMs ensures these captions remain coherent and natural.
To train the LLM, we create a small ``fusing'' dataset with ChatGPT \cite{brown2020language}, and then use it to fine-tune the open-source FlanT5-XL model \cite{chung2022scaling}.


\paragraph{ChatGPT Annotation}\label{Se:chatgpt-annotation}
We leverage the zero-shot capabilities of ChatGPT to generate a ``fusing'' dataset of 20K examples, which is then used to fine-tune an open-source LLM.
By employing this approach, we establish a framework based on open-source models that can generate a large-scale dataset at a reasonable cost.
To produce enriched captions with ChatGPT, we first extract information from the visual experts presented in \Cref{Se:vision-experts}. 
The objects and their corresponding attributes and detected texts are then ordered from left to right based on their bounding boxes, providing basic spatial context.
The exact prompt can be found in the appendix.
We generated 20K examples of such enriched captions from CC, CC12, SBU, and COCO datasets,
which served as training data for the FlanT5-XL model.
\vspace{-6pt}


\paragraph{LLM Fine-tuning}
Using the ``fusing'' dataset we created, we fine-tuned Flan-T5 \cite{chung2022scaling}, a variant of the T5 encoder-decoder model \cite{raffel2020exploring}, that has been extensively fine-tuned on numerous tasks to achieve exceptional performance on instruction-based tasks.
Specifically, we utilized the Flan-T5-XL checkpoint and fine-tuned it to our curated fuse dataset.
During the fine-tuning process, the original caption was concatenated with the output of the visual experts and served as the model input, while the enriched caption was designated as the target. The hyperparameters we used for this section can be found in the appendix.
After this phase, the fine-tuned model can be used to generate an enriched captions dataset.
\vspace{-3pt}


\section{Training a Captioning Model}
\label{subsec:stage2}
We apply \fusecap{}, presented in \Cref{Se:method}, to the COCO, SBU, CC, and CC12 datasets, yielding 12 million augmented image-caption pairs.
Following our data-centric approach, and to demonstrate the effectiveness of this augmented dataset, we use it to train a captioning model based on the BLIP architecture \cite{li2022blip} (\Cref{method_stage2}).
Adopting the training strategy from the original BLIP paper, we first perform vision-language pre-training for $20$ epochs , optimizing three objectives:
\begin{inparaenum}[(a)]
\item \textbf{Image-Text Contrastive Loss (ITC)} that aligns visual and textual features by contrasting between matching and non-matching image-text pairs, 
\item \textbf{Image-Text Matching Loss (ITM)} that classifies image-text pairs as matched or unmatched, and
\item \textbf{Generative Language Modeling Loss (LM)} that generates textual captions from images, which are compared to their ground truth captions.
\end{inparaenum}
Following the pre-training phase, adhering to common practice in VLP \cite{li2022blip, li2023blip, vinvl}, we fine-tune our model for a supplementary $5$ epochs on the \emph{enriched} COCO dataset, utilizing solely LM loss.
To allow the generation of more comprehensive captions, we increase the context length over that used in the original BLIP model from $30$ to $60$ tokens.
The hyperparameters for this section are detailed in the appendix.
\section{Experiments}\label{Se:experiments}
We evaluate two sets of image captions.
The first set consists of the fused captions from the \fusecap{} dataset, described in \Cref{Se:method}. The second set includes captions generated by the trained captioning model discussed in \Cref{subsec:stage2}. 
The objective for both sets is to produce captions that are descriptive and accurate.
Traditionally, image captioning methods are assessed with n-gram-based metrics like BLEU \cite{papineni2002bleu}, CIDEr \cite{vedantam2015cider}, and ROUGE \cite{lin2004rouge}. 
These metrics compare the tested captions to reference captions, assuming the latter represents the ideal image descriptions that the tested captions aim for.
However, as highlighted earlier, captions in existing datasets often fail to provide a comprehensive description of images.
This implies that given our emphasis on enhancing descriptiveness, relying on broad and non-specific reference captions for evaluation would be inappropriate.
Consequently, n-gram-based metrics do not effectively measure or promote this quality \cite{zhu2023chatgpt}. 

We, therefore adopt CLIPScore~\cite{hessel2022clipscore}, a reference-free metric that measures the alignment between textual and visual embeddings generated by a pre-trained CLIP model \cite{radford2021learning}.
Since CLIPScore is not dependent on a reference caption, its score is not restricted by the descriptiveness of the original captions.
Moreover, in terms of accuracy, it has been shown to exhibit a higher correlation with human judgments than n-gram-based metrics \cite{hessel2022clipscore}.

Besides evaluating enriched captions with CLIPScore, we consider the ability to perform image-text retrieval as a pertinent metric for caption evaluation.
Comprehensive captions should inherently serve as distinct image descriptors, thereby improving retrieval precision.
To further assess the quality of captions in the \fusecap{} dataset, we conducted a human evaluation study.

\subsection{FuseCap Dataset}
\label{Se:data-validation}
In this subsection, we evaluate the enriched captions produced by the caption augmentation approach introduced in the \Cref{Se:method}.
In particular, we carry out a qualitative human-study alongside quantitative evaluations.
Examples of enriched captions via FuseCap are showcased in \Cref{fig:generatior_example2}.
\vspace{-20pt}


\begin{table}[b]
  \centering
  \vspace{-9pt}
  \begin{tabular}{clccccc}
  \toprule
    Dataset & Captions & Mean & Voting  \\
    \midrule
    \multirow{2}{*}{COCO} & Original & 76.7 & 31.7\%\\
    & \fusecap{} & \textbf{80.3} & \textbf{67.6\%} \\
    \midrule
    \multirow{2}{*}{SBU} & Original & 71.9 & 32.1\%\\
    & \fusecap{} & \textbf{75.5} & \textbf{60.2\%} \\
    \midrule
    \multirow{2}{*}{CC} & Original & 72.6 & 34.7\%\\
    & \fusecap{} & \textbf{75.4} & \textbf{59.7\%} \\
    \bottomrule
  \end{tabular}
    \caption{\textbf{\fusecap{} data quantitative evaluation.} CLIPScore-based comparison between the original captions of common image-text datasets with our enriched ones. ``Mean'' indicated the mean CLIPScore and ``Voting'' expresses CLIP's preference in a one-vs-one setting. As can be seen, \fusecap{} obtains significantly improved results in both metrics.}
    \label{tab:caption_clips}
\end{table}


\paragraph{Qualitative Evaluation}
We conducted a thorough human evaluation study to assess the ability of our enriched captions to be both descriptive and relevant to the images. Specifically, we randomly sampled 400 pairs from the COCO dataset 
and provided 40 participants with (1) the original caption and (2) the enriched caption.
Our study engaged a pool of random internet users as participants. To minimize biases and ensure an impartial evaluation, they completed the survey unaware of the specific research objectives or goals.
The evaluators were asked the following:
\textit{``Does caption 2 provide an additional meaningful and truthful description of the image compared to caption 1?''}.
As the enriched captions are much more detailed, as can be seen in \Cref{fig:generatior_example2}, the question focuses on whether they are accurate.
Our study results indicate that participants find the enriched captions at least as good as the original ones in $72.9\%$ of the images. This finding highlights the proposed method's effectiveness in enhancing the captions' descriptiveness while preserving alignment and relevancy to the images.

\vspace{-10pt}
\paragraph{Quantitative Evaluation}
We evaluate the enriched captions in comparison to the original captions for each dataset under consideration. To this end, we randomly selected 5000 images from each dataset and report the CLIPScore obtained with both types of captions.
As depicted in \Cref{tab:caption_clips}, the enriched captions generated by our proposed method consistently achieve a higher CLIPScore (Mean) on average by $4.6\%$.
In addition, given an image and two captions (original and enriched), we utilize CLIP to measure which caption is preferred. We evaluate this on the different datasets and summarize it under ``Voting'', which our captions outperform current ones on average by $29.7\%$.
These results demonstrate the effectiveness of our approach in generating enriched captions that better reflect the content of the images.
\vspace{-20pt}

\paragraph{Image-Text Retrieval}\label{Se:image-text-retrieval}
To further demonstrate the descriptiveness and accuracy of the \fusecap{} data, we assess its performance in the image-text retrieval task.
This task involves matching images to text queries and vice versa.
If the enriched data is descriptive and accurate, retrieval performance should improve since the additional details can serve as a discriminative factor in establishing these correspondences.
Our training methodology aligns with the original BLIP model.
After pre-training the model on a large-scale dataset as in \Cref{subsec:stage2}, the model is fine-tuned for image-text retrieval on the COCO training set using both ITC and ITM losses.
For the inference step, we employ the method proposed by \cite{li2021align}, previously integrated into BLIP.
This method involves the selection of $K$ candidates based on feature similarity, followed by their re-ranking using respective ITM values.
We report $R@N$ that corresponds to the accuracy of retrieving the true text/image among the top $N$ retrieved results.
The details of the competing models are provided in \Cref{Se:caption-generation}, with the exception that the versions discussed here have been fine-tuned for the retrieval tasks.
As illustrated in \Cref{coco_retrieval}, the use of fused captions contributes significantly to the enhancement of retrieval performance on the COCO test set\footnote{While we used enriched captions for training and testing BLIP$_\fusecap{}$, for the other baselines we used the original dataset without enrichment.}.
For example, compared to the model trained on corresponding non-enriched data, the $R@1$ score for image-to-text retrieval increased by $22.1\%$, and for text-to-image retrieval, it increased by $34.8\%$.

\begin{table}
  \centering
\begin{tabular}{l *{3}{c@{\hspace{5pt}}} | *{3}{c@{\hspace{5pt}}}}
    \toprule
    & \multicolumn{6}{c}{COCO Retrieval} \\
    & \multicolumn{3}{c}{img $\rightarrow$ text} & \multicolumn{3}{c}{text $\rightarrow$ img} \\
    \cmidrule(lr){2-4} \cmidrule(lr){5-7}
    Model & 
    R\kern-0.1em@\kern-0.1em1 & R\kern-0.1em@\kern-0.1em5 & R\kern-0.1em@\kern-0.1em10 & R\kern-0.1em@\kern-0.1em1 & R\kern-0.1em@\kern-0.1em5 & R\kern-0.1em@\kern-0.1em10 \\
    \midrule
    BLIP$\dag$ & 75.1 & 92.7 & 96.4 & 58.2 & 82.4 & 89.2 \\
    BLIP-L & 82.4 & 95.4 & 97.9 & 65.2 & 86.3 & 91.8 \\
    BLIP2 & 85.4 & 97.0 & 98.5 & 68.3 & 87.7 & 92.6 \\
    \midrule
    BLIP$^*_\fusecap{}$ & \textbf{97.2} & \textbf{99.5} & \textbf{99.9} & \textbf{93.0} & \textbf{97.4} & \textbf{98.3} \\
    \bottomrule
\end{tabular}
\vspace{7pt}
  \caption{\textbf{Image-text retrieval results.} Performance on COCO retrieval (test sets). The ``*'' symbol indicates that the model was trained and tested on our enriched dataset. These results attest that given rich captions, BLIP$_\fusecap{}$ significantly outperforms existing methods, which utilize standard captions.
  }
  \label{coco_retrieval}
  \vspace{-0.3cm}
\end{table}

\subsection{Caption Generation}\label{Se:caption-generation}
We fine-tune the BLIP model that was pre-trained on the complete \fusecap{} dataset for captioning, using the enriched COCO dataset.
We refer to this fine-tuned version as BLIP$_\fusecap{}$.
To assess the effectiveness of the captioner, and, by extension, the \fusecap{} data it is trained on, we thoroughly compare our results to various state-of-the-art captioning models using CLIPScore.
In particular, we consider the following baselines:
\begin{itemize}
\item \textbf{BLIP$\dag$}~\cite{li2022blip}: An original BLIP model, pre-trained and fine-tuned on the same image set as BLIP$_\fusecap{}$, which uses \textbf{original} captions, in contrast to BLIP$_\fusecap{}$ which uses the enriched captions. To guarantee a fair comparison, we set configuration parameters identical to those implemented in BLIP$_\fusecap{}$. 
\item \textbf{BLIP-L}: A large version of BLIP, which was pre-trained on a dataset comprising 129M images prior to the captioning fine-tuning.
Among all models presented in the original BLIP paper, this model achieves superior captioning results.
\item \textbf{BLIP2-G-OPT$_{2.7}$}~\cite{li2023blip}: A state-of-the-art captioning model that integrates a large frozen vision backbone (ViT G) along with a frozen LLM (OPT \cite{zhang2022opt}), trained on the datasets considered for BLIP-L.
\item \textbf{OFA}~\cite{wang2022ofa}: A state-of-the-art image captioning method trained on unimodal and multimodal pre-training tasks and fine-tuned for captioning.
\item \textbf{GIT}~\cite{git}: A state-of-the-art vision-language model, with an image encoder and text decoder architecture. It is scaled up in terms of both pre-training data and model size, and it is fine-tuned for image captioning.

\item \textbf{Prismer}~\cite{liu2023prismer}:  A Vision-Language Model with Multi-Modal Experts such as an object detector and an OCR model. Prismer adopts a strategy that could be seen as dual to ours, integrating vision experts into the model architecture, as opposed to leveraging them to enrich the training data.
\end{itemize}

\begin{table}
  \centering
\begin{tabular}{lcc|cc}
    \toprule
    Model & Images & Parameters & Val & Test \\
    \midrule
    BLIP$\dag$     & 12M  & 247M & 75.2  & 75.3 \\
    BLIP-L         & 129M  & 470M & 76.1  & 76.0 \\
    OFA            & 20M  & 470M & 76.6  & 76.4 \\
    GIT            & 800M  & 700M & 77.1  & 77.0 \\
    BLIP2-G-OPT$_{2.7}$ & 129M  & 3.8B & 77.8  & 77.5 \\
    Prismer        & 13M  & 1.6B & 76.7   & 76.7 \\ 
    \midrule
    BLIP$_\fusecap{}$        & 12M  & 247M & \textbf{78.3}   & \textbf{78.5} \\
    \bottomrule
    \end{tabular}
    \caption{
    \textbf{Image captioning results.} CLIPScore of leading models and our approach on the COCO captions dataset. Our model outperforms much larger models that have pre-trained with significantly more training data. 
    }
  \label{caption_clips}

    \end{table}

Building on the methodology outlined in \Cref{Se:experiments}, we evaluate the performance of our models using the mean CLIPScore metric.
As shown in \Cref{caption_clips}, our model not only outperforms the BLIP$\dag$ model by $4.3\%$ but also surpasses the best performing among other models by $1.3\%$.
This improvement over models with a considerably larger parameter count that have been trained on significantly more image-caption pairs underscores the effectiveness of our data centric approach.
Furthermore, as illustrated in \Cref{fig:generatior_example}, our approach exhibits clear superiority in the generation of captions, producing descriptions with greater semantic detail compared to those generated by competing models.
Our results surpassing Prismer indicate that for generating comprehensive image captions, leveraging vision experts to enrich data can be more beneficial than incorporating them directly into the model architecture.
Additional examples showcasing the performance and capabilities of our model can be found in the supplementary material.

\textbf{Image-Text Retrieval}
To further assess the quality of the generated captions, we once again utilize the image-text retrieval task, following the method detailed in \Cref{Se:image-text-retrieval}.
This time, however, we consider captions not from the ground truth COCO or the ``ground truth'' enriched COCO datasets.
Instead, they are produced by the BLIP$\dag$ and BLIP$_\fusecap{}$ caption generators.
Captions from BLIP$\dag$ and BLIP$_\fusecap{}$ are assessed using retrieval models fine-tuned on original and enriched captions, respectively,
both of these retrieval models are the ones discussed in \Cref{Se:image-text-retrieval}.
\Cref{generators_coco_retrieval} shows that by using the captions from the captioners instead of the ground truth captions, BLIP$_\fusecap{}$ retains its superiority over BLIP$\dag$ in both image-to-text and text-to-image retrievals.  Impressively, using the BLIP$_\fusecap{}$ captions, the retrieval performance is on par with results achieved using the ground truth enriched captions in \Cref{coco_retrieval}.
In contrast, BLIP$\dag$ exhibits a significant performance drop when compared to its results with the corresponding ground truth captions.

\begin{table}
  \centering
\begin{tabular}{l *{3}{c@{\hspace{5pt}}} | *{3}{c@{\hspace{5pt}}}}
    \toprule
    & \multicolumn{6}{c}{COCO Retrieval} \\
    & \multicolumn{3}{c}{img $\rightarrow$ text} & \multicolumn{3}{c}{text $\rightarrow$ img} \\
    \cmidrule(lr){2-4} \cmidrule(lr){5-7}
    Model & 
    R\kern-0.1em@\kern-0.1em1 & R\kern-0.1em@\kern-0.1em5 & R\kern-0.1em@\kern-0.1em10 & R\kern-0.1em@\kern-0.1em1 & R\kern-0.1em@\kern-0.1em5 & R\kern-0.1em@\kern-0.1em10 \\
    \midrule
    \vspace{-0.3em}
    \multirow{1.5}{*}{BLIP$\dag$} & 56.3 & 83.0 & 90.3 & 54.5 & 81.2 & 88.7 \\
    \vspace{-0.3em}
      & \footnotesize{\color{OrangeRed}-18.8\%} & \footnotesize{\color{OrangeRed}-9.7\%} & \footnotesize{\color{OrangeRed}-6.1\%} & \footnotesize{\color{OrangeRed}-3.7\%} & \footnotesize{\color{OrangeRed}-1.2\%} & \footnotesize{\color{OrangeRed}-0.5\%} \\
    \midrule
     \vspace{-0.3em}
    \multirow{1.5}{*}{BLIP$^*_\fusecap{}$} & \textbf{95.0} & \textbf{98.8} & \textbf{99.2} & \textbf{94.5} & \textbf{98.7} & \textbf{99.3} \\
     \vspace{-0.3em}
     & \footnotesize{\color{OrangeRed}-2.2\%} & \footnotesize{\color{OrangeRed}-0.7\%} & \footnotesize{\color{OrangeRed}-0.7\%} & \footnotesize{\color{OliveGreen}+1.5\%} & \footnotesize{\color{OliveGreen}+1.3\%} & \footnotesize{\color{OliveGreen}+1\%} \\
    \bottomrule
\end{tabular}
  \caption{\textbf{Image-text retrieval results using generated captions.}
  Performance on COCO retrieval (test sets) using captions generated by the models presented in \Cref{subsec:stage2} rather than ground truth captions.
  The models used for evaluation are the ones referenced in \Cref{Se:image-text-retrieval}.
  The highlighted percentages indicate the performance difference when compared to the use of corresponding ground truth captions in \Cref{coco_retrieval}. 
  Notably, BLIP$_\fusecap{}$ captions excel over BLIP$\dag$, a trend consistent with ground truth captions. 
  Moreover, BLIP$_\fusecap{}$ maintains a performance parity with the ground truth datasets, while BLIP$\dag$ shows a noticeable decline in performance.
  }
  \label{generators_coco_retrieval}
  \vspace{-0.4cm}
\end{table}

\subsection{Large-Scale Data Influence}\label{sec:pretrain_significance}
Typical VLP frameworks involve pre-training vision-language models on image-caption pairs before fine-tuning them on downstream tasks.
Given this structure, one might argue that the generation of enriched captions could serve as an additional downstream task.
This would suggest using the standard captions dataset for pre-training, followed by fine-tuning on a smaller enriched dataset.
Yet, as mentioned in \Cref{Se:caption-generation}, our approach with the BLIP$_\fusecap{}$ model deviates from this by pre-training on enriched datasets instead of the conventional ones. In this section, we evaluate the impact of this deviation.

We compare the performance of models that are pre-trained on both the standard and enriched datasets.
Both models are subsequently fine-tuned using the enriched COCO dataset.
Unlike the evaluation method detailed in Section \ref{Se:experiments}, which is reference-free, this section employs conventional, reference-based evaluation metrics, as both models are fine-tuned and tested using the same dataset.
\Cref{ablation_table} reveals that the model pre-trained on the enriched captions outperforms its counterpart, which was pre-trained on the original captions, across all metrics when both are fine-tuned on the augmented COCO dataset.
This outcome underscores the importance of using large-scale enriched datasets for pre-training and validates our automated dataset creation approach.
Interestingly, the model that is pre-trained on the enriched captions and fine-tuned using the standard COCO dataset performs comparably to the model pre-trained on the original captions and fine-tuned on the same dataset.



\section{Limitations}

In the human evaluation study (\Cref{Se:data-validation}), a subset of participants favored the original caption.
This may be due to our fusing process occasionally missing inter-element dependencies, as seen in \Cref{fig:generatior_example2}, where the cat's position between the man and the laptop was not captured.
Future research might integrate finer visual details, like segmentation, and refine the LLM fuser accordingly.
\vspace{-8pt}

\paragraph{Ethical considerations} \fusecap{} dataset was constructed using FlanT5-XL model \cite{chung2022scaling}, which was trained on unfiltered data potentially laden with explicit content or inherent biases. Consequently, the proposed dataset may replicate biases from the original model.
\section{Conclusions}

\begin{table}
  \centering
  \begin{tabular}{c|c|ccc}
    \toprule
    \makecell{Pre-training\\Data}  & \makecell{Fine-tune\\+Test Data} & B\kern-0.1em@\kern-0.1em4 & CIDEr &  SPICE \\
    \midrule
    Standard  & Standard     &  37.8 &  126.5  &  22.9 \\
    \fusecap{}  & Standard     &  \textbf{38.4} &  \textbf{128.7}  &  \textbf{23.0} \\
    \midrule
    Standard  & \fusecap{}     &  35.4 &  111.4  &  25.0  \\
    \fusecap{}& \fusecap{}     &  \textbf{37.3} &  \textbf{123.1}  &  \textbf{26.8} \\
    \bottomrule
  \end{tabular}
  \caption{\textbf{Influence of Large-Scale Data.} 
    Pre-training with the \fusecap{} dataset significantly outperforms pre-training on standard data when fine-tuned for enriched caption generation.
    This underscores the value of the large-scale \fusecap{} dataset and its automated creation approach.
    Additionally, when fine-tuned for standard captioning, the \fusecap{} pre-trained model is comparable to, if not slightly better than, the standard data pre-trained model.
  }
  \label{ablation_table}
  \vspace{-0.3cm}
\end{table}

In this paper, we address the problem of generating highly descriptive and detailed image captions. We observe that existing state-of-the-art methods produce short and often oversimplified captions that fail to capture the intricate details in images. We hypothesize that this is due to datasets limitation, \textit{i.e.,} existing training data composed of concise captions. Thus, the tendency to provide such captions is distilled into the trained models, regardless of their architecture or training method. Therefore, we introduce a novel and generic data-centric automated strategy to enrich existing image captions, termed \fusecap{}.
Specifically, this approach harnesses visual experts to extract meaningful information from images and an LLM to fuse such data into the existing captions, yielding enriched ones. We apply \fusecap{} to different widespread datasets and generate 12M image-enriched caption pairs.
The augmented image captions quality is evaluated qualitatively using a human evaluation survey, and quantitatively, using different evaluation methods.
Finally, we demonstrate the effectiveness of the enriched data by utilizing it to train an image captioning model, which outperforms significantly larger state-of-the-art methods.
We posit that our research highlights the marked potential of LLMs in enhancing powerful data-centric approaches in computer vision.

{\small
\bibliographystyle{ieee_fullname}
\bibliography{egbib}
}

\clearpage
\pagebreak
\appendix

\section{\fusecap{} Fused Captions}
\begin{figure}[b]
    \centering
    \includegraphics[width=2.1\columnwidth]{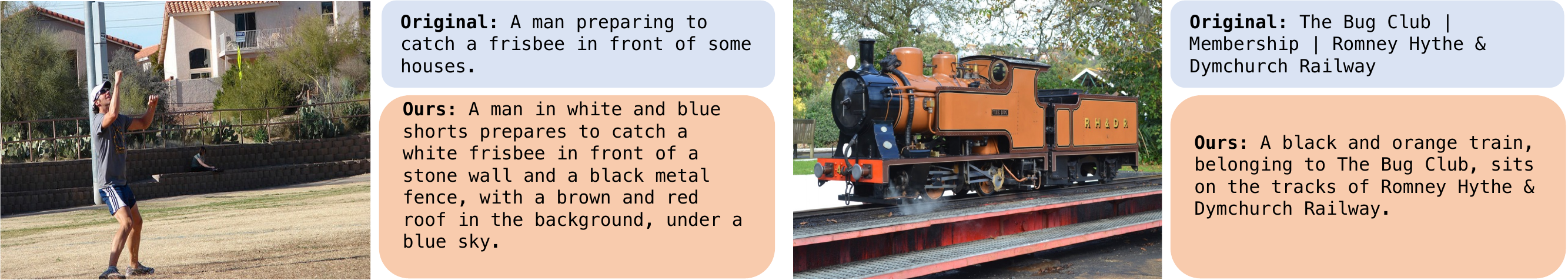}
    \label{fig:supp_fusecap_examples}
    \vspace{1cm}
\end{figure}

In this section, we present supplementary details about the newly proposed fused dataset presented in Section \ref{Se:method}.
\Cref{fig:supp_fusecap_examples} presents examples of this enriched data, adding to the ones shown in Figure \ref{fig:generatior_example2}.


\subsection{Visual Experts Implementation}
We utilize the Visual Experts discussed in Section 
\ref{Se:vision-experts}, in the following manner,
\begin{itemize}
    \item \textbf{Object detection} We consider objects as valid detections if they surpass a pre-determined detection confidence threshold of 0.7.
    \item \textbf{Attribute Detection} We incorporate attributes to each valid predicted object if the attribute confidence surpasses a 0.2 threshold.
    \item \textbf{OCR}: We use CRAFT and Parseq with default inference parameters \cite{baek2019character, bautista2022parseq}. The text recognized is attributed to the object that has the smallest bounding box encompassing it.
\end{itemize}

\subsection{Training Set Generation for LLM Fuser with ChatGPT.}
As outlined in Section \ref{Se:llm-fuser}, we harness the zero-shot capabilities of ChatGPT to generate a compact dataset encompassing 20,000 examples. This data is subsequently used to fine-tune an open-source Large Language Model (LLM).
The prompt provided to ChatGPT is as follows:

\noindent\makebox[\linewidth]{\rule{0.48\textwidth}{0.4pt}}

"A caption of an image is given: \textit{original caption}. \\
The following objects are detected in the image from left to right: \\
\quad A $a^1_1$, ..., $a^{k-1}_1$ and $a^{k}_1$ $o_1$ [with the following text: $t_1$]. \\
\quad \vdots \\
\quad A $a^1_n$, ..., $a^{k_{n}-1}_n$ and $a^{k_{n}}_n$ $o_n$ [with the following text: $t_n$]. \\
Write a comprehensive and concise caption of the scene using the objects detected."\\
\noindent\makebox[\linewidth]{\rule{0.48\textwidth}{0.4pt}}

\vfill

\newpage
\noindent We denote $\{o_i\}_{i=1}^{N}$ as the set of objects detected, $\{a^{j}_{i}\}_{i=1}^{N}$ as the attributes related to each object, and $\{t_i\}_{i=1}^{N}$ as texts related to each object.


\section{Training Settings}
\subsection{LLM Fuser}
Our LLM Fuser is a fine-tuned FlanT5-XL \cite{chung2022scaling} model. We used the huggingface library to fine-tune it on a single NVIDIA A40 GPU. We trained the model for $4,000$ optimization steps, with batch size of $32$, a learning rate of $5\cdot10^{-5}$, and a linear scheduling strategy. We limit the source and target length to 100 and 200 tokens, respectively. 

\subsection{Caption Generator}
\Cref{fig:supp_captioner_examples} offers further examples, supplementing those found in \Cref{fig:generatior_example} and providing additional instances of the model's outputs.
These examples further emphasize the ability of the captioning model to generate semantically rich captions.

\textbf{Training.}
The pre-training and subsequent fine-tuning of the captioning model, described in Section \ref{Se:experiments}, was performed on eight NVIDIA A100 GPUs.
For setting the pre-training hyperparameters, we followed the approach outlined in the original BLIP implementation \cite{li2022blip}, except for maximum caption length, batch size, and initial learning rate.
We deviated from the original model's batch size to accommodate the increased token length used in our implementation.
To maintain stability during pre-training with a smaller batch size, we reduced the initial learning rate.
The batch size used for pre-training was $400$ ($50$ per GPU).
The initial learning rate used was $6 \cdot 10 ^{-5}$.

\textbf{Maximum Caption Length.}
The original BLIP model imposes a training and inference limit of $30$ tokens per caption. With our enrichment process, however, caption lengths tend to be longer and beyond this original threshold.
Accordingly, we have increased the maximum caption length to 60 tokens.
This limit is maintained through both the pre-training and fine-tuning stages of the caption generator, as well as during the actual generation of captions.

\onecolumn
\renewcommand{\LTcaptype}{figure}
\begin{longtable}{c}
  \captionsetup{type=figure} 
  \includegraphics[width=1\linewidth]{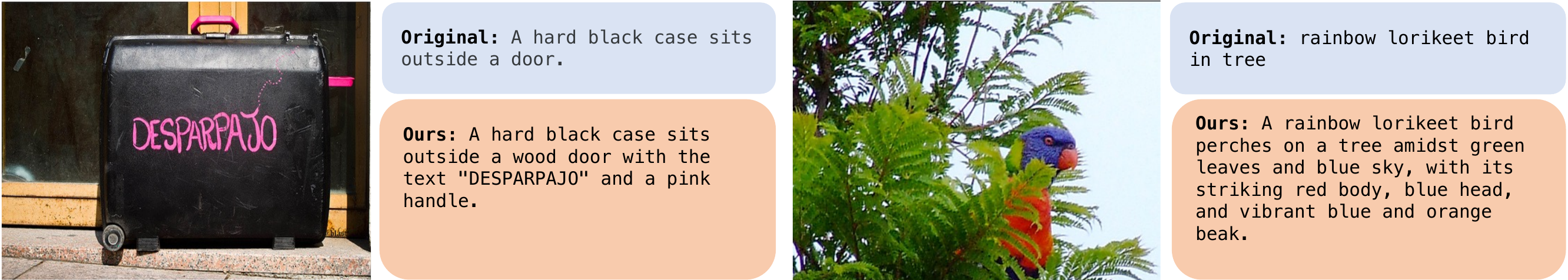}\\
  \includegraphics[width=1\linewidth]{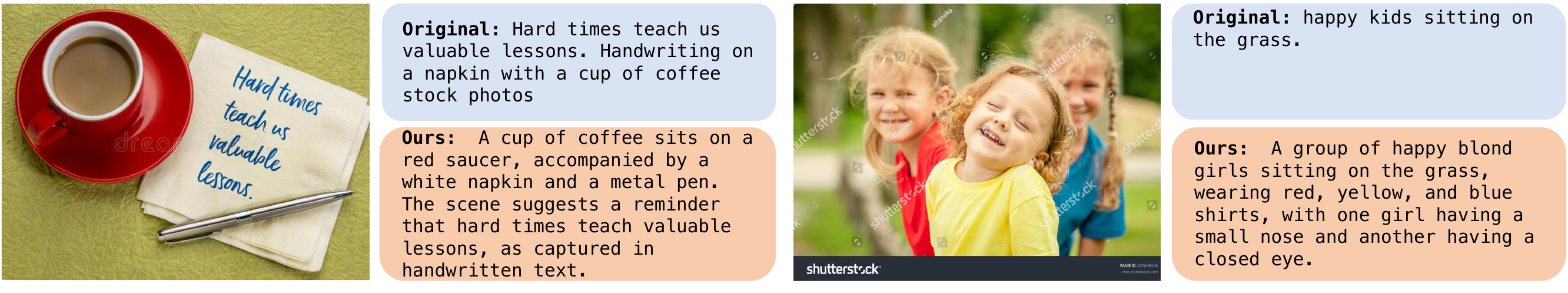}\\
  \includegraphics[width=1\linewidth]{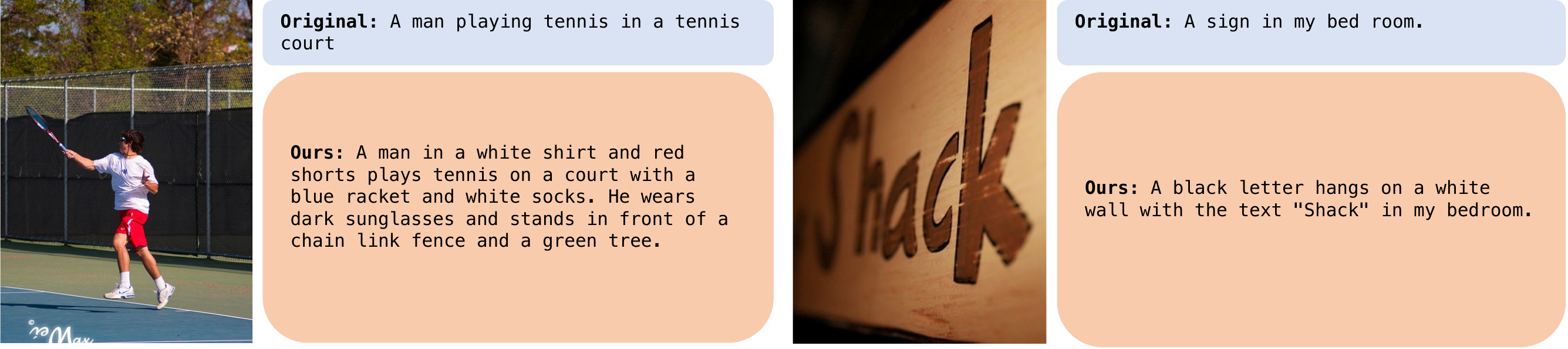}\\
  \caption{Examples of the proposed \fusecap{} enriched-captions dataset.}
\end{longtable}

\renewcommand{\LTcaptype}{figure}
\begin{longtable}{c}
  \includegraphics[width=0.9\linewidth]{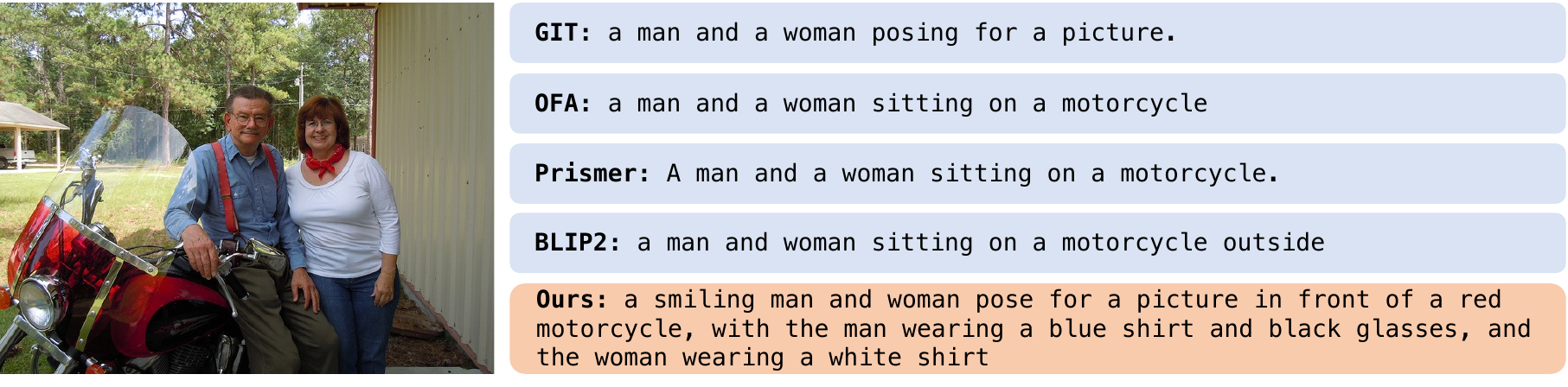}\\
  \includegraphics[width=0.9\linewidth]{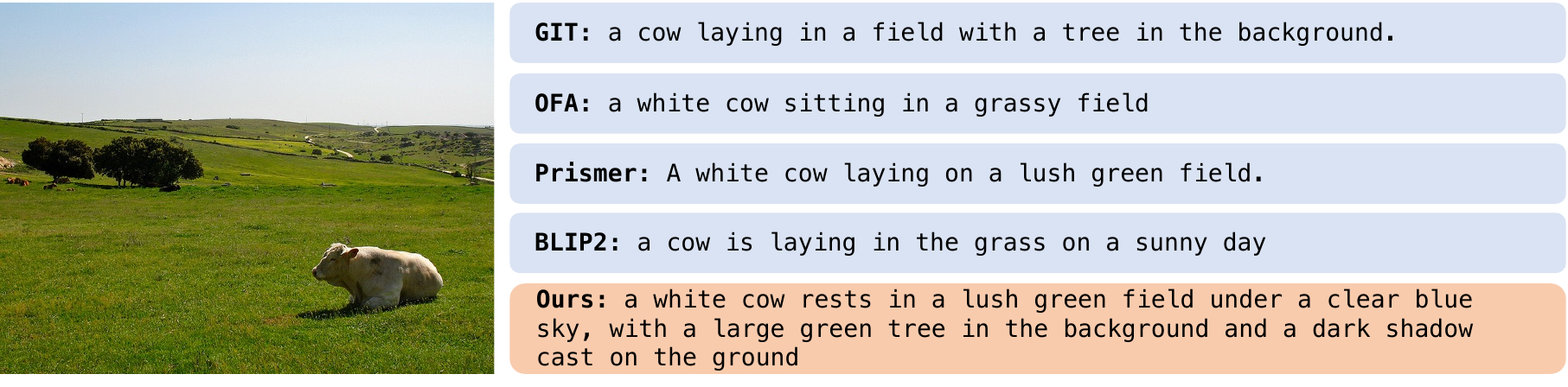}\\
  \includegraphics[width=0.9\linewidth]{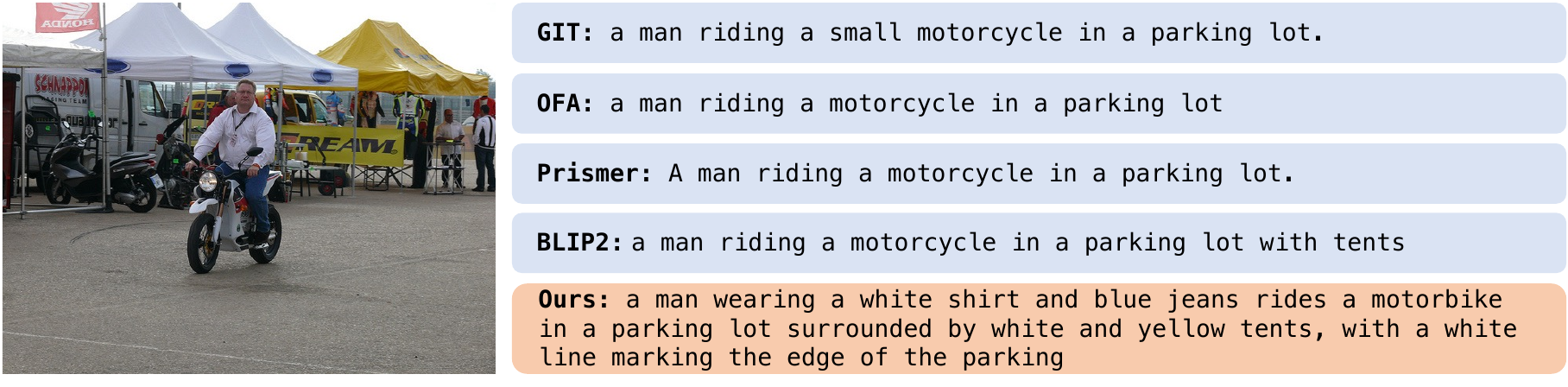}\\
  \includegraphics[width=0.9\linewidth]{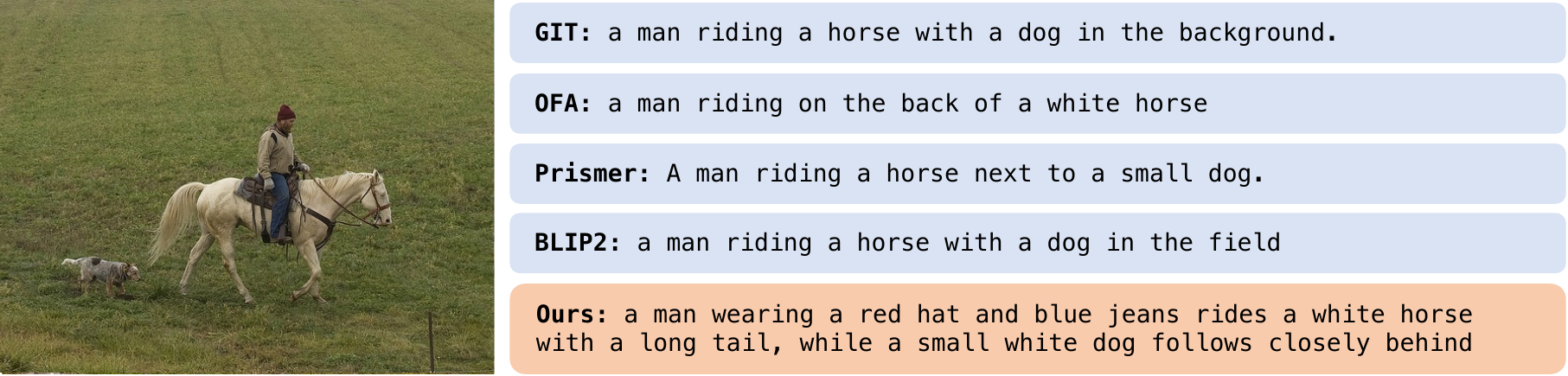}\\
  \includegraphics[width=0.9\linewidth]{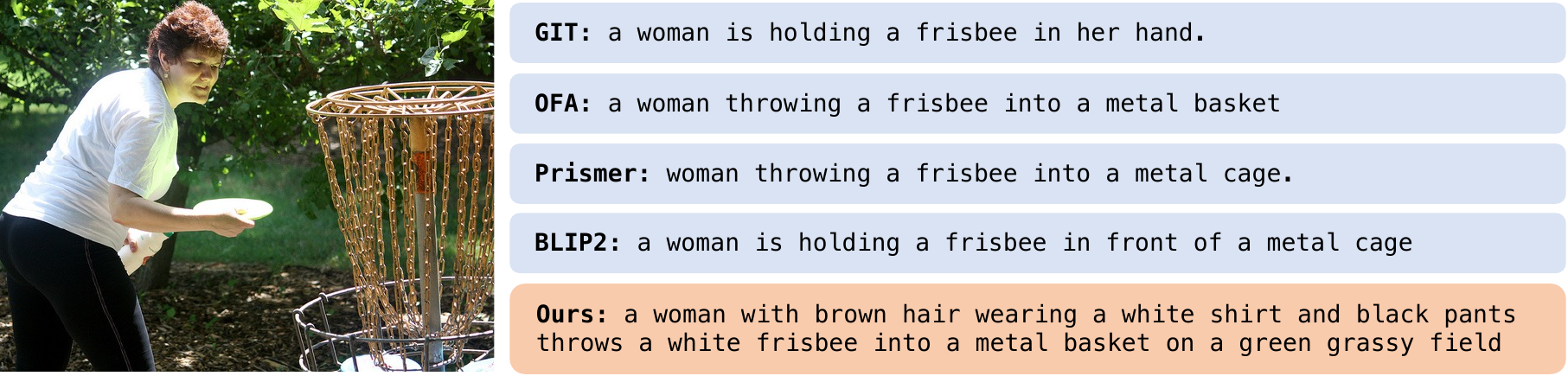}\\
  \includegraphics[width=0.9\linewidth]{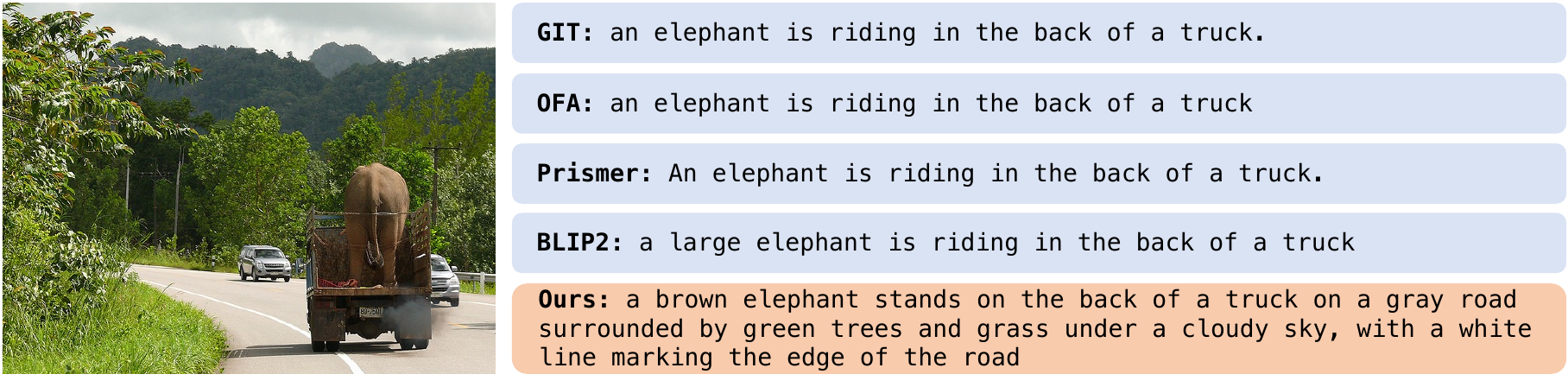}\\
  \includegraphics[width=0.9\linewidth]{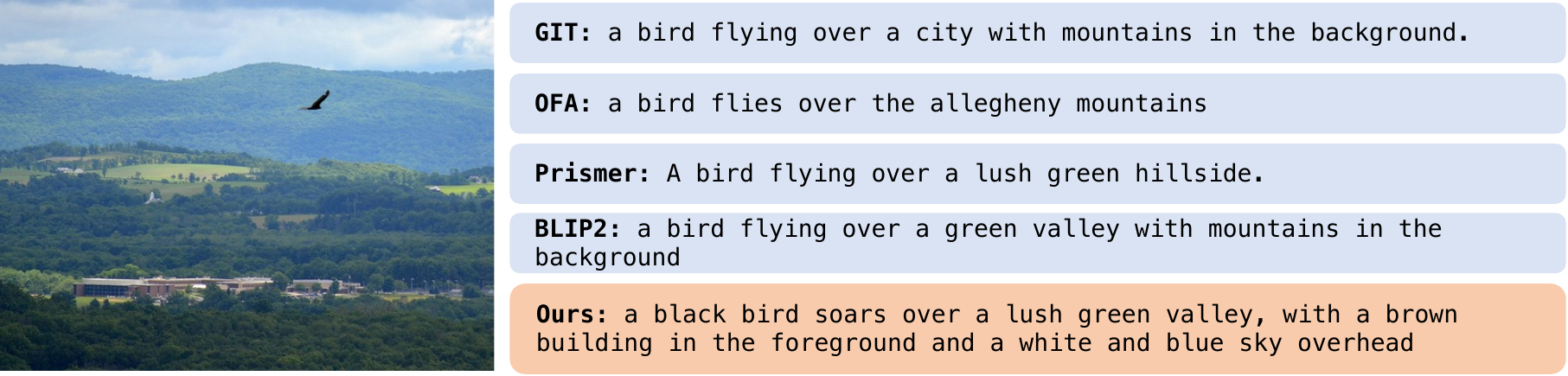}\\
  \includegraphics[width=0.9\linewidth]{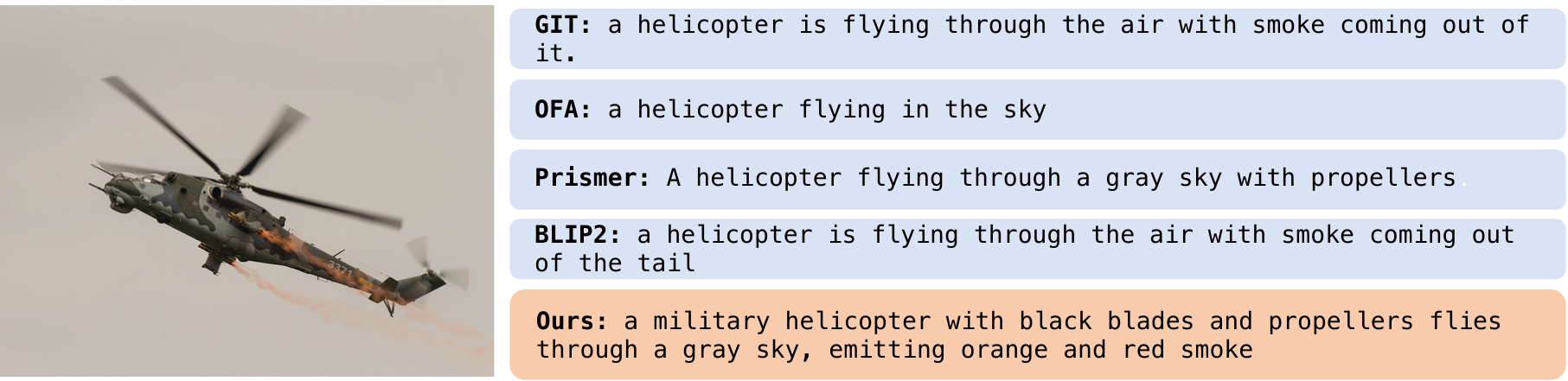}\\
  \includegraphics[width=0.75\linewidth]{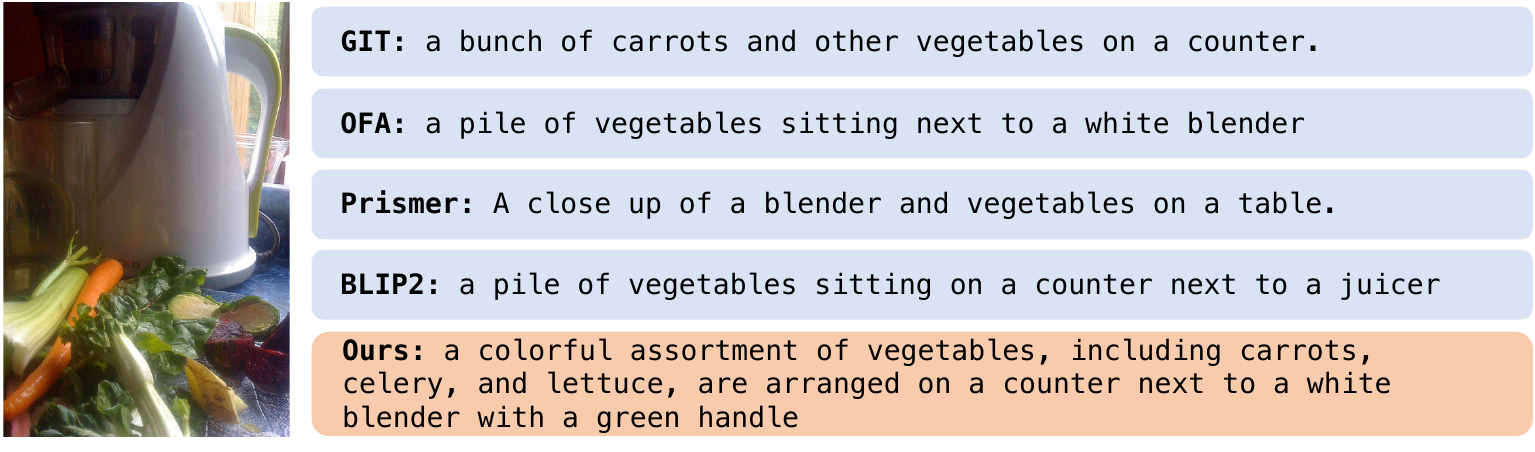}\\
  \includegraphics[width=0.75\linewidth]{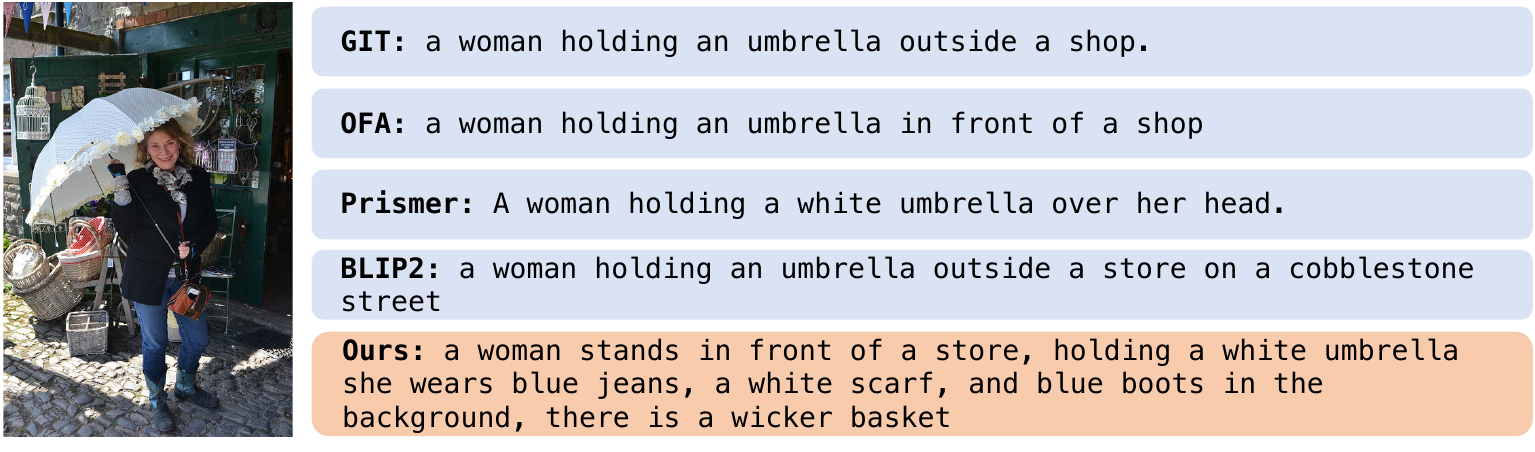}\\
  \caption{Comparative illustration of captions generated by our BLIP$_\fusecap{}$ and other top-performing models.
  \label{fig:supp_captioner_examples}}\\
\end{longtable}

\end{document}